\documentclass{article}

\PassOptionsToPackage{numbers,compress}{natbib}

\usepackage[dblblindworkshop, final]{neurips_2025}
\workshoptitle{The First Workshop on Generative and Protective AI for Content Creation}

\usepackage[utf8]{inputenc} 
\usepackage[T1]{fontenc}    
\usepackage{hyperref}       
\usepackage{url}            
\usepackage{booktabs}       
\usepackage{amsfonts}       
\usepackage{nicefrac}       
\usepackage{microtype}      
\usepackage{xcolor}         
\usepackage{graphicx}
\usepackage{colortbl}
\usepackage{multirow}
\usepackage{amsmath, amssymb}
\usepackage{subcaption}

\title{Evaluating the Evaluators: Metrics for Compositional Text-to-Image Generation}

\author{%
    Seyed Amir Kasaei\textsuperscript{1} \\
    \texttt{a.kasaei@me.com} \\
    \And
    Ali Aghayari\textsuperscript{1} \\
    \texttt{ali.aghayari@hotmail.com} \\
    \And
    Arash Marioriyad\textsuperscript{1} \\
    \texttt{arashmarioriyad@gmail.com} \\
    \And
    Niki Sepasian\textsuperscript{1} \\
    \texttt{sepasian.niki@gmail.com} \\
    \And
    MohammadAmin Fazli\textsuperscript{1} \\
    \texttt{fazli@sharif.edu} \\
    \And
    Mahdieh Soleymani Baghshah\textsuperscript{1} \\
    \texttt{soleymani@sharif.edu} \\
    \And
    Mohammad Hossein Rohban\textsuperscript{1} \\
    \texttt{rohban@sharif.edu} \\
    \AND
    \normalfont\textsuperscript{1}Department of Computer Engineering,\\
    Sharif University of Technology \\
}

\usepackage{pifont}

\begin{document}
\maketitle

\begin{abstract}
Text–image generation has advanced rapidly, but assessing whether outputs truly capture the objects, attributes, and relations described in prompts remains a central challenge. Evaluation in this space relies heavily on automated metrics, yet these are often adopted by convention or popularity rather than validated against human judgment.
Because evaluation and reported progress in the field depend directly on these metrics, it is critical to understand how well they reflect human preferences. To address this, we present a broad study of widely used metrics for compositional text–image evaluation. Our analysis goes beyond simple correlation, examining their behavior across diverse compositional challenges and comparing how different metric families align with human judgments. The results show that no single metric performs consistently across tasks: performance varies with the type of compositional problem. Notably, VQA-based metrics, though popular, are not uniformly superior, while certain embedding-based metrics prove stronger in specific cases. Image-only metrics, as expected, contribute little to compositional evaluation, as they are designed for perceptual quality rather than alignment.
These findings underscore the importance of careful and transparent metric selection, both for trustworthy evaluation and for their use as reward models in generation. Project page is available at \href{https://amirkasaei.com/eval-the-evals/}{this URL}.
\end{abstract}

\section{Introduction}

Recent advances in multi-modal generative models, such as Stable Diffusion \cite{rombach2022high, podell2023sdxl, esser2024scaling} and DALL-E \cite{ramesh2022hierarchical}, have made it possible to generate high-quality, natural, and diverse images from textual descriptions at low cost and on a large scale. Alongside this progress, a parallel line of research has emerged around a central question: \textit{how can we faithfully assess the alignment between text and image, particularly in terms of compositional alignment?}

Text–image compositional alignment refers to how well the objects, attributes, and relations described in a prompt are reflected in the generated image. A variety of metrics have been introduced to evaluate this alignment. \emph{Content-based metrics}, such as VQAScore \cite{lin2024evaluating}, TIFA \cite{hu2023tifa}, DA Score \cite{singh2023divide}, DSG \cite{cho2023davidsonian}, and B-VQA \cite{huang2023t2i}, assess compositional properties through structured queries to the image. \emph{Embedding-based metrics}, including CLIPScore \cite{hessel2021clipscore}, PickScore \cite{kirstain2023pickscore}, HPS \cite{wu2023hps}, and ImageReward \cite{xu2024imagereward}, measure alignment via text–image similarity in a shared representation space or by using models trained on human preference data. \emph{Image-only metrics}, such as CLIP-IQA \cite{wang2023exploring} and Aesthetic Score \cite{schuhmann2022laion}, instead focus on perceptual quality and visual appeal, complementing alignment-based evaluation. Together, these metrics determine how models are evaluated and compared, and much of the field’s reported progress is defined through their outcomes. Selecting appropriate metrics is therefore essential. 


In addition to their role as evaluation tools, text–image alignment metrics have increasingly served as reward signals to enhance compositional generation in diffusion models through both inference-time and reinforcement learning strategies. For instance, ReNO \cite{eyring2024reno} leverages multiple reward models, including CLIPScore \cite{hessel2021clipscore}, ImageReward \cite{xu2024imagereward}, PickScore \cite{kirstain2023pickscore}, and HPSv2 \cite{wu2023hps}, to guide gradient-based optimization of initial noise during inference, improving both faithfulness and aesthetics. In contrast, ImageSelect \cite{karthik2023if} adopts a best-of-$N$ sampling strategy, generating multiple candidate images from different initial noises and selecting the one with the highest \emph{ImageReward} \cite{xu2024imagereward} score. On the training side, reinforcement learning methods like DPOK \cite{fan2023dpok} fine-tune the diffusion model by maximizing ImageReward \cite{xu2024imagereward}, using online policy gradient to strengthen compositional alignment.

Despite their widespread use in both evaluation and generation, there has been no thorough and comparative analysis of how well these metrics reflect human judgment. To address this gap, this work presents a comprehensive evaluation of $12$ text–image compositional alignment metrics on $2400$ generated text–image samples spanning $8$ compositional categories \cite{t2iplus}, examining how closely each metric corresponds to human assessments. 
\section{Text-image Compositional Alignment}



\subsection{Evaluation Metrics}
A range of metrics have been proposed for evaluating text–image alignment, each targeting different aspects of the correspondence. They can be grouped into three categories: (1) \emph{embedding-based}, which rely on representations or preference models; (2) \emph{content-based}, which use structured reasoning to assess compositional properties; and (3) \emph{image-only}, which measure perceptual quality independently of text.

\paragraph{Embedding-based Metrics}
Embedding-based metrics evaluate alignment by comparing text–image representations in a shared multimodal space or by leveraging models trained on human preferences. A common baseline is \emph{CLIPScore} \cite{hessel2021clipscore}, which measures cosine similarity between CLIP embeddings. Preference-supervised variants include \emph{HPS} \cite{wu2023hps}, which fine-tunes CLIP on human comparisons, and \emph{PickScore} \cite{kirstain2023pickscore}, which learns from pairwise preference judgments. \emph{BLIP} \cite{li2023blip2} follows the embedding-similarity approach, comparing captions generated from images with the input text. Extending this idea, \emph{ImageReward} \cite{xu2024imagereward} adds a reward head trained on ranked human preference data, capturing both textual relevance and perceptual quality.



\paragraph{Content-based (VQA-based) Metrics}
VQA-based metrics assess compositional alignment by casting text–image consistency as a question answering task. Questions derived from the prompt are posed to a pretrained VQA model, with scores based on the correctness of its responses. \emph{VQAScore} \cite{lin2024evaluating} generates yes/no questions from the text, while \emph{TIFA} \cite{hu2023tifa} uses structured templates to cover objects, attributes, and relations. Variants target specific aspects: \emph{DA Score} \cite{singh2023divide} asks entity–attribute questions to test binding, \emph{DSG} \cite{cho2023davidsonian} converts the text into a scene graph to verify entities and relations, and \emph{B-VQA} \cite{huang2023t2i} decomposes the text into object–attribute pairs, querying each with BLIP-VQA and combining the probabilities. 


\paragraph{Image-only Metrics}
Image-only metrics assess perceptual quality independently of the prompt, providing complementary signals of realism and aesthetics. \emph{CLIP-IQA} \cite{wang2023exploring} predicts image quality by regressing CLIP embeddings against human quality annotations, while the \emph{Aesthetic Score} \cite{schuhmann2022laion} estimates aesthetic value from large-scale human ratings.

\subsection{Benchmarks}
Text–image compositional alignment benchmarks, such as T2I-CompBench++ \cite{t2iplus}, group prompts into categories that reflect major alignment challenges. These include: \emph{entity existence} (all entities appear without omission or hallucination), \emph{attribute binding} (objects match specified properties like color or shape), \emph{spatial relations} (2D and 3D object arrangements, e.g., “a cube on a sphere” or “a ball inside a box”), \emph{non-spatial relations} (functional interactions, e.g., “a man holding a guitar”), and \emph{numeracy} (accurate object counts, e.g., “five birds on a branch”). Together, these categories provide a structured and fine-grained basis for evaluating compositional alignment.


\section{Experiments and Results}

\subsection{Experimental Setting}
\label{sec:experiment_setting}
Our analysis is based on T2I-CompBench++ \cite{t2iplus}, which provides curated prompts across attributes (color, shape, texture), spatial relations (2D and 3D), non-spatial relations, complex prompts, and numeracy. Each prompt is paired with images from multiple text-to-image models and annotated with human evaluation scores. All resources (prompts, images, and scores) come from the benchmark; our contribution is to analyze how evaluation metrics align with these annotations using outputs from SD v1.4, SD v2, Structured Diffusion \cite{Feng2022TrainingFreeSD}, Composable Diffusion \cite{liu2022compositional}, Attend-and-Excite \cite{Chefer2023AttendandExciteAS}, and GORS \cite{huang2023t2i}.


We evaluate five embedding-based metrics (PickScore \citep{kirstain2023pickscore}, CLIPScore \citep{hessel2021clipscore}, HPS \citep{wu2023hps}, ImageReward \citep{xu2024imagereward}, BLIP-2 \citep{li2023blip2}), two image-only metrics (CLIP-IQA \citep{wang2023exploring}, Aesthetic Score \citep{schuhmann2022laion}), and five VQA-based metrics (B-VQA \citep{huang2023t2i}, DA Score \citep{singh2023divide}, TIFA \citep{hu2023tifa}, DSG \citep{cho2023davidsonian}, VQA Score \citep{lin2024evaluating}), covering embedding similarity, perceptual quality, and VQA-style reasoning.

\subsection{Correlation Analysis of Evaluation Metrics}
We assess the reliability of reward models by correlating their scores on 
T2I-CompBench++ generations with human evaluations \ref{sec:experiment_setting}. Spearman correlations, reported in Table~\ref{tab:correlation_metrics_spearman},vserve as the main measure, while Pearson and Kendall results are provided in Appendix~\ref{app:corr_study} for completeness.


\paragraph{Per-Category Breakdown of Correlation Results}
Table~\ref{tab:correlation_metrics_spearman} highlights that the strongest correlations differ substantially across categories, indicating that \emph{no single metric dominates overall}. In the attribute group, DA Score leads on color (TIFA second), while ImageReward ranks highest on shape and texture (DA Score second). For relational cases, VQA Score performs best on 2D spatial (HPS second), whereas DSG leads in 3D spatial (HPS and BLIP-2 second). Non-spatial relations are best captured by HPS, followed by ImageReward. In complex prompts, VQA Score shows the strongest alignment, with TIFA second, and in numeracy, TIFA ranks first with ImageReward next. \emph{Across all categories, image-only metrics (CLIP-IQA, Aesthetic) remain consistently weak}, underscoring their limited value for compositional alignment.


\begin{table*}[ht]
    \centering
    \caption{Spearman correlation of evaluation metrics with human scores across compositional categories on T2I-CompBench++. The highest value in each category is shown in bold, and the second-highest is underlined.}
    \resizebox{\textwidth}{!}{
        \begin{tabular}{lcccccccc}
            \toprule
            \textbf{Metric} & \textbf{Color} & \textbf{Shape} & \textbf{Texture} & \textbf{2D Spatial} & \textbf{Non-Spatial} & \textbf{Complex} & \textbf{3D Spatial} & \textbf{Numeracy} \\
            \midrule
            CLIP \citep{hessel2021clipscore} & 0.282 & 0.291 & 0.535 & 0.369 & 0.439 & 0.276 & 0.315 & 0.223 \\
            PickScore \citep{kirstain2023pickscore} & 0.263 & 0.270 & 0.516 & 0.299 & 0.432 & 0.167 & 0.139 & 0.337 \\
            HPS \citep{wu2023hps} & 0.219 & 0.440 & 0.601 & \underline{0.410} & \textbf{0.535} & 0.270 & \underline{0.416} & 0.471 \\
            ImageReward \citep{xu2024imagereward} & 0.580 & \textbf{0.520} & \textbf{0.734} & 0.394 & \underline{0.512} & 0.424 & 0.401 & \underline{0.484} \\
            BLIP2 \citep{li2023blip2} & 0.250 & 0.287 & 0.546 & 0.369 & 0.353 & 0.235 & \underline{0.416} & 0.366 \\
            Aesthetic \citep{schuhmann2022laion} & 0.056 & 0.195 & 0.078 & 0.136 & 0.061 & 0.051 & 0.123 & 0.036 \\
            CLIP-IQA \citep{wang2023exploring} & 0.092 & 0.078 & -0.001 & 0.088 & 0.082 & 0.027 & 0.098 & 0.068 \\
            B-VQA \citep{huang2023t2i} & 0.610 & 0.388 & 0.690 & 0.255 & 0.371 & 0.372 & 0.330 & 0.444 \\
            DA Score \citep{singh2023divide} & \textbf{0.772} & \underline{0.463} & \underline{0.711} & 0.318 & 0.453 & 0.488 & 0.297 & 0.462 \\
            TIFA \citep{hu2023tifa} & \underline{0.684} & 0.336 & 0.423 & 0.311 & 0.351 & \underline{0.519} & 0.195 & \textbf{0.526} \\
            DSG \citep{cho2023davidsonian} & 0.599 & 0.388 & 0.628 & 0.328 & 0.470 & 0.411 & \textbf{0.427} & 0.469 \\
            VQA Score \citep{lin2024evaluating} & 0.678 & 0.405 & 0.701 & \textbf{0.533} & 0.495 & \textbf{0.638} & 0.339 & 0.473 \\
            \bottomrule
        \end{tabular}
    }
    
    \label{tab:correlation_metrics_spearman}
\end{table*}

\paragraph{Broader Insights on Metric Performance}
Several broader insights emerge from these results. First, \emph{no single metric achieves strong and consistent correlation across all compositional categories}, indicating that reliance on a single signal is insufficient. Second, despite its widespread use \citep{rombach2022high,nichol2021glide,ruiz2023dreambooth,brooks2023instructpix2pix,kumari2023multi,kang2023scaling,chefer2023attend,podell2023sdxl,chen2023pixartalpha,li2023gligen,nguyen2024swiftbrush}, \emph{CLIP never ranks among the top metrics}, underscoring its limitations as a standalone measure. Third, embedding-based metrics, particularly ImageReward and HPS, frequently appear among the strongest. Fourth, while VQA-based metrics are competitive, \emph{they are not uniformly superior and are occasionally outperformed by embedding-based approaches}. Finally, image-only metrics such as CLIP-IQA and Aesthetic remain consistently weak, as expected since they do not assess text–image alignment.


\subsection{Beyond Correlation: Regression Analysis of Metrics}
To complement the correlation study, we ran a regression analysis using human scores as the target and all metric outputs as predictors. A separate linear model was fit for each compositional category, with coefficients shown in Table~\ref{tab:regression_coefficients}. These coefficients reflect each metric’s joint contribution, revealing shifts in importance across categories. Embedding-based metrics such as HPS and PickScore show strong positive contributions, while ImageReward and VQA-based metrics (DA Score, VQA Score, TIFA) also play key roles. In contrast, CLIP-IQA and Aesthetic often receive negligible or negative coefficients. Overall, embedding-based and VQA-based metrics provide complementary signals, and while the exact top metric can differ from correlation results, the same families tend to dominate—highlighting that no individual metric proves universally strong across categories.

\begin{table*}[ht]
    \centering
    \caption{Regression coefficients for predicting human scores on T2I-CompBench++ categories. Highest values are bold, second-highest underlined.}
    \resizebox{\textwidth}{!}{
        \begin{tabular}{lcccccccc}
            \toprule
            \textbf{Metric} & \textbf{Color} & \textbf{Shape} & \textbf{Texture} & \textbf{2D Spatial} & \textbf{Non-Spatial} & \textbf{Complex} & \textbf{3D Spatial} & \textbf{Numeracy} \\
            \midrule
            CLIP \citep{hessel2021clipscore} & -0.290 & 0.000 & -0.148 & 0.000 & \textbf{0.649} & 0.000 & -0.444 & 0.000 \\
            PickScore \citep{kirstain2023pickscore} & \textbf{2.467} & 0.000 & \textbf{1.564} & 0.000 & 0.000 & 0.000 & -2.801 & 0.000 \\
            HPS \citep{wu2023hps} & -0.097 & \textbf{0.761} & \underline{0.475} & \textbf{1.143} & \underline{0.629} & 0.000 & \textbf{2.048} & \textbf{1.277} \\
            ImageReward \citep{xu2024imagereward} & 0.193 & \underline{0.334} & 0.300 & 0.197 & -0.095 & \underline{0.223} & 0.008 & 0.053 \\
            BLIP2 \citep{li2023blip2} & -0.420 & 0.000 & 0.264 & 0.000 & 0.594 & 0.000 & \underline{0.510} & 0.000 \\
            Aesthetic \citep{schuhmann2022laion} & 0.229 & 0.000 & 0.260 & 0.000 & 0.023 & 0.000 & -0.194 & 0.000 \\
            CLIP-IQA \citep{wang2023exploring} & -0.017 & 0.000 & 0.014 & 0.016 & 0.000 & 0.000 & 0.100 & 0.000 \\
            B-VQA \citep{huang2023t2i} & -0.092 & -0.063 & 0.059 & -0.264 & 0.035 & 0.000 & 0.071 & 0.059 \\
            DA Score \citep{singh2023divide} & \underline{0.754} & 0.132 & 0.236 & \underline{0.428} & 0.000 & 0.000 & -0.223 & 0.051 \\
            TIFA \citep{hu2023tifa} & 0.211 & 0.028 & -0.015 & 0.003 & 0.061 & 0.214 & -0.002 & \underline{0.204} \\
            DSG \citep{cho2023davidsonian} & -0.043 & 0.068 & 0.101 & 0.000 & 0.023 & 0.000 & 0.161 & -0.045 \\
            VQA Score \citep{lin2024evaluating} & 0.112 & 0.082 & 0.064 & 0.319 & 0.068 & \textbf{0.246} & -0.003 & 0.092 \\
            \bottomrule
        \end{tabular}
    }
    
    \label{tab:regression_coefficients}
\end{table*}

\subsection{Distribution Patterns of Metric Scores}
\begin{figure*}[ht]
    \centering
    \includegraphics[width=\textwidth]{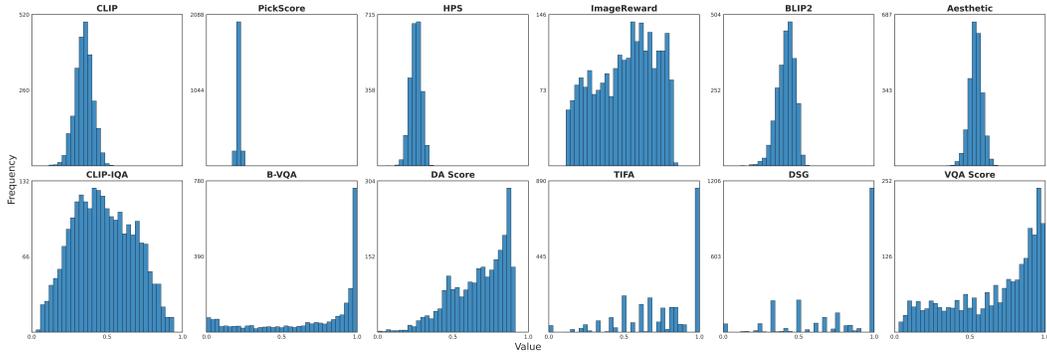}
    \caption{Value distributions of all analyzed metrics over T2I-CompBench++ generations (bin counts normalized by frequency).}
    \label{fig:metric_value_ranges}
\end{figure*}

Different families of evaluation metrics exhibit distinct distributional behaviors (Figure~\ref{fig:metric_value_ranges}).
Embedding-based metrics (CLIPScore, PickScore, HPS, ImageReward, BLIP-2) tend to produce mid-range scores. CLIPScore, HPS, and BLIP-2 are concentrated around values of 0.25–0.5, while ImageReward spans a broader range (0.15–0.8).
Image-only metrics behave differently: Aesthetic is narrowly concentrated around 0.5–0.6, whereas CLIP-IQA spans a much wider range of 0.1–0.9.
In contrast, VQA-based metrics (B-VQA, DA Score, TIFA, DSG, VQA Score) are strongly right-skewed and often saturate near 1.0, reflecting their quasi-binary nature.
These patterns reveal two major concerns for evaluation. First, certain embedding-based metrics such as CLIP have a restricted value range, with many samples clustered around mid-level scores, which makes it difficult to distinguish quality differences and undermines their usefulness as evaluation measures. Second, VQA-based metrics are biased toward high values, often saturating near the upper bound, which reduces their ability to separate stronger candidates.

\section{Conclusion}
We studied how well current evaluation metrics capture human judgment in compositional text–to–image generation. Our analysis showed that no single metric is consistently reliable across categories: embedding-based methods (e.g., ImageReward, HPS) and VQA-based metrics (e.g., DA Score, VQA Score) each contribute, but with varying strengths. Regression confirmed that their contributions shift when combined, while distributional patterns revealed limitations such as compressed mid-range scores in embeddings and saturation in VQA metrics. These results highlight the need for combining complementary metrics and for more faithful evaluation practices to guide progress in the field.

\newpage
{\small
\bibliographystyle{abbrvnat} 
\bibliography{ref}

@article{nichol2021glide,
  title={Glide: Towards photorealistic image generation and editing with text-guided diffusion models},
  author={Nichol, Alex and Dhariwal, Prafulla and Ramesh, Aditya and Shyam, Pranav and Mishkin, Pamela and McGrew, Bob and Sutskever, Ilya and Chen, Mark},
  journal={arXiv preprint arXiv:2112.10741},
  year={2021}
}

@article{ramesh2022hierarchical,
  title={Hierarchical text-conditional image generation with clip latents},
  author={Ramesh, Aditya and Dhariwal, Prafulla and Nichol, Alex and Chu, Casey and Chen, Mark},
  journal={arXiv preprint arXiv:2204.06125},
  volume={1},
  number={2},
  pages={3},
  year={2022}
}

@inproceedings{rombach2022high,
  title={High-resolution image synthesis with latent diffusion models},
  author={Rombach, Robin and Blattmann, Andreas and Lorenz, Dominik and Esser, Patrick and Ommer, Bj{\"o}rn},
  booktitle={Proceedings of the IEEE/CVF conference on computer vision and pattern recognition},
  pages={10684--10695},
  year={2022}
}

@article{podell2023sdxl,
  title={Sdxl: Improving latent diffusion models for high-resolution image synthesis},
  author={Podell, Dustin and English, Zion and Lacey, Kyle and Blattmann, Andreas and Dockhorn, Tim and M{\"u}ller, Jonas and Penna, Joe and Rombach, Robin},
  journal={arXiv preprint arXiv:2307.01952},
  year={2023}
}

@inproceedings{esser2024scaling,
  title={Scaling rectified flow transformers for high-resolution image synthesis},
  author={Esser, Patrick and Kulal, Sumith and Blattmann, Andreas and Entezari, Rahim and M{\"u}ller, Jonas and Saini, Harry and Levi, Yam and Lorenz, Dominik and Sauer, Axel and Boesel, Frederic and others},
  booktitle={Forty-first International Conference on Machine Learning},
  year={2024}
}

@misc{chen2023pixartalpha,
      title={PixArt-$\alpha$: Fast Training of Diffusion Transformer for Photorealistic Text-to-Image Synthesis}, 
      author={Junsong Chen and Jincheng Yu and Chongjian Ge and Lewei Yao and Enze Xie and Yue Wu and Zhongdao Wang and James Kwok and Ping Luo and Huchuan Lu and Zhenguo Li},
      year={2023},
      eprint={2310.00426},
      archivePrefix={arXiv},
      primaryClass={cs.CV}
}

@article{huang2023t2i,
  title={T2i-compbench: A comprehensive benchmark for open-world compositional text-to-image generation},
  author={Huang, Kaiyi and Sun, Kaiyue and Xie, Enze and Li, Zhenguo and Liu, Xihui},
  journal={Advances in Neural Information Processing Systems},
  volume={36},
  pages={78723--78747},
  year={2023}
}

@article{t2iplus,
  title={T2i-compbench++: An enhanced and comprehensive benchmark for compositional text-to-image generation},
  author={Huang, Kaiyi and Duan, Chengqi and Sun, Kaiyue and Xie, Enze and Li, Zhenguo and Liu, Xihui},
  journal={IEEE Transactions on Pattern Analysis and Machine Intelligence},
  year={2025},
  publisher={IEEE}
}

@inproceedings{wang2023exploring,
  title={Exploring clip for assessing the look and feel of images},
  author={Wang, Jianyi and Chan, Kelvin CK and Loy, Chen Change},
  booktitle={Proceedings of the AAAI conference on artificial intelligence},
  volume={37},
  number={2},
  pages={2555--2563},
  year={2023}
}

@article{schuhmann2022laion,
  title={Laion-5b: An open large-scale dataset for training next generation image-text models},
  author={Schuhmann, Christoph and Beaumont, Romain and Vencu, Richard and Gordon, Cade and Wightman, Ross and Cherti, Mehdi and Coombes, Theo and Katta, Aarush and Mullis, Clayton and Wortsman, Mitchell and others},
  journal={Advances in neural information processing systems},
  volume={35},
  pages={25278--25294},
  year={2022}
}

@article{hessel2021clipscore,
  title={Clipscore: A reference-free evaluation metric for image captioning},
  author={Hessel, Jack and Holtzman, Ari and Forbes, Maxwell and Bras, Ronan Le and Choi, Yejin},
  journal={arXiv preprint arXiv:2104.08718},
  year={2021}
}

@inproceedings{li2023blip2,
  title={Blip-2: Bootstrapping language-image pre-training with frozen image encoders and large language models},
  author={Li, Junnan and Li, Dongxu and Savarese, Silvio and Hoi, Steven},
  booktitle={International conference on machine learning},
  pages={19730--19742},
  year={2023},
  organization={PMLR}
}

@article{kirstain2023pickscore,
  title={Pick-a-pic: An open dataset of user preferences for text-to-image generation},
  author={Kirstain, Yuval and Polyak, Adam and Singer, Uriel and Matiana, Shahbuland and Penna, Joe and Levy, Omer},
  journal={Advances in Neural Information Processing Systems},
  volume={36},
  pages={36652--36663},
  year={2023}
}

@inproceedings{wu2023hps,
  title={Human preference score: Better aligning text-to-image models with human preference},
  author={Wu, Xiaoshi and Sun, Keqiang and Zhu, Feng and Zhao, Rui and Li, Hongsheng},
  booktitle={Proceedings of the IEEE/CVF International Conference on Computer Vision},
  pages={2096--2105},
  year={2023}
}

@article{xu2024imagereward,
  title={Imagereward: Learning and evaluating human preferences for text-to-image generation},
  author={Xu, Jiazheng and Liu, Xiao and Wu, Yuchen and Tong, Yuxuan and Li, Qinkai and Ding, Ming and Tang, Jie and Dong, Yuxiao},
  journal={Advances in Neural Information Processing Systems},
  volume={36},
  year={2024}
}

@inproceedings{hu2023tifa,
  title={Tifa: Accurate and interpretable text-to-image faithfulness evaluation with question answering},
  author={Hu, Yushi and Liu, Benlin and Kasai, Jungo and Wang, Yizhong and Ostendorf, Mari and Krishna, Ranjay and Smith, Noah A},
  booktitle={Proceedings of the IEEE/CVF International Conference on Computer Vision},
  pages={20406--20417},
  year={2023}
}

@inproceedings{lin2024evaluating,
  title={Evaluating text-to-visual generation with image-to-text generation},
  author={Lin, Zhiqiu and Pathak, Deepak and Li, Baiqi and Li, Jiayao and Xia, Xide and Neubig, Graham and Zhang, Pengchuan and Ramanan, Deva},
  booktitle={European Conference on Computer Vision},
  pages={366--384},
  year={2024},
  organization={Springer}
}

@article{cho2023davidsonian,
  title={Davidsonian scene graph: Improving reliability in fine-grained evaluation for text-image generation},
  author={Cho, Jaemin and Hu, Yushi and Garg, Roopal and Anderson, Peter and Krishna, Ranjay and Baldridge, Jason and Bansal, Mohit and Pont-Tuset, Jordi and Wang, Su},
  journal={arXiv preprint arXiv:2310.18235},
  year={2023}
}

@article{singh2023divide,
  title={Divide, evaluate, and refine: Evaluating and improving text-to-image alignment with iterative vqa feedback},
  author={Singh, Jaskirat and Zheng, Liang},
  journal={Advances in Neural Information Processing Systems},
  volume={36},
  pages={70799--70811},
  year={2023}
}

@inproceedings{ruiz2023dreambooth,
  title={Dreambooth: Fine tuning text-to-image diffusion models for subject-driven generation},
  author={Ruiz, Nataniel and Li, Yuanzhen and Jampani, Varun and Pritch, Yael and Rubinstein, Michael and Aberman, Kfir},
  booktitle={Proceedings of the IEEE/CVF conference on computer vision and pattern recognition},
  pages={22500--22510},
  year={2023}
}

@article{chefer2023attend,
  title={Attend-and-excite: Attention-based semantic guidance for text-to-image diffusion models},
  author={Chefer, Hila and Alaluf, Yuval and Vinker, Yael and Wolf, Lior and Cohen-Or, Daniel},
  journal={ACM Transactions on Graphics (TOG)},
  volume={42},
  number={4},
  pages={1--10},
  year={2023},
  publisher={ACM New York, NY, USA}
}

@inproceedings{liu2022compositional,
  title={Compositional visual generation with composable diffusion models},
  author={Liu, Nan and Li, Shuang and Du, Yilun and Torralba, Antonio and Tenenbaum, Joshua B},
  booktitle={European Conference on Computer Vision},
  pages={423--439},
  year={2022},
  organization={Springer}
}

@article{karthik2023if,
  title={If at First You Don't Succeed, Try, Try Again: Faithful Diffusion-based Text-to-Image Generation by Selection},
  author={Karthik, Shyamgopal and Roth, Karsten and Mancini, Massimiliano and Akata, Zeynep},
  journal={arXiv preprint arXiv:2305.13308},
  year={2023}
}

@article{eyring2024reno,
  title={ReNO: Enhancing One-step Text-to-Image Models through Reward-based Noise Optimization}, 
  author={Luca Eyring and Shyamgopal Karthik and Karsten Roth and Alexey Dosovitskiy and Zeynep Akata},
  journal={Neural Information Processing Systems (NeurIPS)},
  year={2024}
}

@article{Chefer2023AttendandExciteAS,
  title={Attend-and-Excite: Attention-Based Semantic Guidance for Text-to-Image Diffusion Models},
  author={Hila Chefer and Yuval Alaluf and Yael Vinker and Lior Wolf and Daniel Cohen-Or},
  journal={ACM Transactions on Graphics (TOG)},
  year={2023},
  volume={42},
  pages={1 - 10},
  url={https://api.semanticscholar.org/CorpusID:256416326}
}

@inproceedings{
Feng2022TrainingFreeSD,
title={Training-Free Structured Diffusion Guidance for Compositional Text-to-Image Synthesis},
author={Weixi Feng and Xuehai He and Tsu-Jui Fu and Varun Jampani and Arjun Reddy Akula and Pradyumna Narayana and Sugato Basu and Xin Eric Wang and William Yang Wang},
booktitle={The Eleventh International Conference on Learning Representations },
year={2023},
url={https://openreview.net/forum?id=PUIqjT4rzq7}
}

@inproceedings{brooks2023instructpix2pix,
  title={Instructpix2pix: Learning to follow image editing instructions},
  author={Brooks, Tim and Holynski, Aleksander and Efros, Alexei A},
  booktitle={Proceedings of the IEEE/CVF conference on computer vision and pattern recognition},
  pages={18392--18402},
  year={2023}
}

@inproceedings{kumari2023multi,
  title={Multi-concept customization of text-to-image diffusion},
  author={Kumari, Nupur and Zhang, Bingliang and Zhang, Richard and Shechtman, Eli and Zhu, Jun-Yan},
  booktitle={Proceedings of the IEEE/CVF conference on computer vision and pattern recognition},
  pages={1931--1941},
  year={2023}
}

@inproceedings{kang2023scaling,
  title={Scaling up gans for text-to-image synthesis},
  author={Kang, Minguk and Zhu, Jun-Yan and Zhang, Richard and Park, Jaesik and Shechtman, Eli and Paris, Sylvain and Park, Taesung},
  booktitle={Proceedings of the IEEE/CVF conference on computer vision and pattern recognition},
  pages={10124--10134},
  year={2023}
}

@inproceedings{nguyen2024swiftbrush,
  title={Swiftbrush: One-step text-to-image diffusion model with variational score distillation},
  author={Nguyen, Thuan Hoang and Tran, Anh},
  booktitle={Proceedings of the IEEE/CVF Conference on Computer Vision and Pattern Recognition},
  pages={7807--7816},
  year={2024}
}

@inproceedings{li2023gligen,
  title={Gligen: Open-set grounded text-to-image generation},
  author={Li, Yuheng and Liu, Haotian and Wu, Qingyang and Mu, Fangzhou and Yang, Jianwei and Gao, Jianfeng and Li, Chunyuan and Lee, Yong Jae},
  booktitle={Proceedings of the IEEE/CVF conference on computer vision and pattern recognition},
  pages={22511--22521},
  year={2023}
}

@article{fan2023dpok,
  title={Dpok: Reinforcement learning for fine-tuning text-to-image diffusion models},
  author={Fan, Ying and Watkins, Olivia and Du, Yuqing and Liu, Hao and Ryu, Moonkyung and Boutilier, Craig and Abbeel, Pieter and Ghavamzadeh, Mohammad and Lee, Kangwook and Lee, Kimin},
  journal={Advances in Neural Information Processing Systems},
  volume={36},
  pages={79858--79885},
  year={2023}
}
}
\newpage
\appendix

{\Large\textbf{Appendix}}

\section{Correlation Study}\label{app:corr_study}

In addition to Spearman correlation (Table~\ref{tab:correlation_metrics_spearman} in the main text), we also report Pearson and Kendall correlation results in Appendix Tables~\ref{tab:correlation_metrics_pearson} and \ref{tab:correlation_metrics_kendall}. These complementary analyses confirm the same overall trend: no single metric consistently dominates across all categories, and the strongest performers vary depending on the type of compositional challenge. To further summarize these findings, Table~\ref{tab:top3_metrics_count} shows the frequency with which each metric appears among the top three in a category. This highlights the relative robustness of VQA Score and ImageReward, which each occur in six categories, followed by DA Score and HPS with four each. Importantly, the final set of consistently strong metrics includes two VQA-based methods (VQA Score, DA Score) and two embedding-based methods (ImageReward, HPS), suggesting that both types of evaluation are necessary for capturing the full spectrum of compositional alignment.

\begin{table*}[ht]
    \centering
    \caption{Pearson correlation of evaluation metrics with human scores across compositional categories on T2I-CompBench++. The highest value in each category is shown in bold, and the second-highest is underlined.}
    \resizebox{\textwidth}{!}{
        \begin{tabular}{lcccccccc}
            \toprule
            \textbf{Metric} & \textbf{Color} & \textbf{Shape} & \textbf{Texture} & \textbf{2D Spatial} & \textbf{Non-Spatial} & \textbf{Complex} & \textbf{3D Spatial} & \textbf{Numeracy} \\
            \midrule
            CLIP \citep{hessel2021clipscore} & 0.275 & 0.276 & 0.530 & 0.345 & 0.651 & 0.349 & 0.306 & 0.210 \\
            PickScore \citep{kirstain2023pickscore} & 0.227 & 0.263 & 0.492 & 0.314 & 0.558 & 0.205 & 0.175 & 0.376 \\
            HPS \citep{wu2023hps} & 0.240 & 0.457 & 0.555 & \underline{0.431} & \textbf{0.623} & 0.355 & \textbf{0.460} & \underline{0.495} \\
            ImageReward \citep{xu2024imagereward} & 0.585 & \textbf{0.555} & \textbf{0.737} & 0.403 & 0.543 & 0.496 & 0.405 & 0.482 \\
            BLIP2 \citep{li2023blip2} & 0.259 & 0.333 & 0.541 & 0.344 & \underline{0.620} & 0.299 & 0.412 & 0.364 \\
            Aesthetic \citep{schuhmann2022laion} & 0.078 & 0.195 & 0.068 & 0.152 & 0.095 & 0.077 & 0.097 & 0.061 \\
            CLIP-IQA \citep{wang2023exploring} & 0.114 & 0.080 & 0.019 & 0.070 & 0.015 & 0.023 & 0.108 & 0.018 \\
            B-VQA \citep{huang2023t2i} & 0.621 & 0.348 & 0.644 & 0.237 & 0.462 & 0.426 & 0.335 & 0.460 \\
            DA Score \citep{singh2023divide} & \textbf{0.793} & \underline{0.464} & \underline{0.674} & 0.324 & 0.519 & 0.513 & 0.298 & 0.485 \\
            TIFA \citep{hu2023tifa} & \underline{0.685} & 0.364 & 0.415 & 0.298 & 0.451 & \underline{0.552} & 0.212 & \textbf{0.530} \\
            DSG \citep{cho2023davidsonian} & 0.569 & 0.392 & 0.598 & 0.332 & 0.507 & 0.355 & \underline{0.441} & 0.450 \\
            VQA Score \citep{lin2024evaluating} & 0.682 & 0.409 & 0.617 & \textbf{0.452} & 0.542 & \textbf{0.613} & 0.348 & 0.487 \\
            \bottomrule
        \end{tabular}
    }
    
    \label{tab:correlation_metrics_pearson}
\end{table*}

\begin{table*}[ht]
    \centering
    \caption{Kendall’s $\tau$ correlation of evaluation metrics with human scores across compositional categories on T2I-CompBench++. The highest value in each category is shown in bold, and the second-highest is underlined.}
    \resizebox{\textwidth}{!}{
        \begin{tabular}{lcccccccc}
            \toprule
            \textbf{Metric} & \textbf{Color} & \textbf{Shape} & \textbf{Texture} & \textbf{2D Spatial} & \textbf{Non-Spatial} & \textbf{Complex} & \textbf{3D Spatial} & \textbf{Numeracy} \\
            \midrule
            CLIP \citep{hessel2021clipscore} & 0.208 & 0.211 & 0.392 & 0.287 & 0.347 & 0.201 & 0.224 & 0.154 \\
            PickScore \citep{kirstain2023pickscore} & 0.193 & 0.192 & 0.373 & 0.229 & 0.341 & 0.122 & 0.100 & 0.241 \\
            HPS \citep{wu2023hps} & 0.157 & 0.326 & 0.441 & \underline{0.315} & \textbf{0.428} & 0.201 & \underline{0.305} & 0.346 \\
            ImageReward \citep{xu2024imagereward} & 0.434 & \textbf{0.388} & \textbf{0.549} & 0.310 & \underline{0.408} & 0.313 & 0.294 & 0.349 \\
            BLIP2 \citep{li2023blip2} & 0.179 & 0.203 & 0.389 & 0.286 & 0.280 & 0.170 & 0.303 & 0.264 \\
            Aesthetic \citep{schuhmann2022laion} & 0.039 & 0.138 & 0.054 & 0.104 & 0.047 & 0.037 & 0.083 & 0.026 \\
            CLIP-IQA \citep{wang2023exploring} & 0.065 & 0.055 & -0.002 & 0.068 & 0.063 & 0.018 & 0.068 & 0.045 \\
            B-VQA \citep{huang2023t2i} & 0.456 & 0.279 & 0.512 & 0.195 & 0.293 & 0.267 & 0.231 & 0.322 \\
            DA Score \citep{singh2023divide} & \textbf{0.603} & 0.337 & \underline{0.534} & 0.247 & 0.357 & 0.364 & 0.206 & 0.347 \\
            TIFA \citep{hu2023tifa} & \underline{0.559} & 0.246 & 0.329 & 0.266 & 0.292 & \underline{0.405} & 0.155 & \textbf{0.400} \\
            DSG \citep{cho2023davidsonian} & 0.499 & \underline{0.303} & 0.503 & 0.292 & \underline{0.408} & 0.325 & \textbf{0.355} & \underline{0.363} \\
            VQA Score \citep{lin2024evaluating} & 0.512 & 0.292 & 0.516 & \textbf{0.422} & 0.390 & \textbf{0.481} & 0.243 & 0.352 \\
            \bottomrule
        \end{tabular}
    }
    
    \label{tab:correlation_metrics_kendall}
\end{table*}

\begin{table*}[ht]
    \centering
    \caption{Top-3 presence of each metric across various compositional categories. A \checkmark{} indicates the metric is among the top 3 in that category. The last column shows the total number of categories where the metric appears in the top 3 based on spearman correlation.}
    \resizebox{\textwidth}{!}{
        \begin{tabular}{l|cccccccc|c}
            \toprule
            \textbf{Metric} & \textbf{Color} & \textbf{Shape} & \textbf{Texture} & \textbf{2D Spatial} & \textbf{Non-Spatial} & \textbf{Complex} & \textbf{3D Spatial} & \textbf{Numeracy} & \textbf{Total} \\
            \midrule
            CLIP \citep{hessel2021clipscore} &  &  &  &  &  &  &  &  & 0 \\
            \midrule
            PickScore \citep{kirstain2023pickscore} &  &  &  &  &  &  &  &  & 0 \\
            \midrule
            HPS \citep{wu2023hps} &  & \checkmark &  & \checkmark & \checkmark &  & \checkmark & & \textbf{4} \\
            \midrule
            ImageReward \citep{xu2024imagereward} &  & \checkmark & \checkmark & \checkmark & \checkmark &  & \checkmark & \checkmark & \textbf{6} \\
            \midrule
            BLIP2 \citep{li2023blip2} &  &  &  &  &  &  & \checkmark &  & 1 \\
            \midrule
            Aesthetic \citep{schuhmann2022laion} &  &  &  &  &  &  &  &  & 0 \\
            \midrule
            CLIP-IQA \citep{wang2023exploring} &  &  &  &  &  &  &  &  & 0 \\
            \midrule
            B-VQA \citep{huang2023t2i} &  &  &  &  &  &  &  &  & 0 \\
            \midrule
            DA Score \citep{singh2023divide} & \checkmark & \checkmark & \checkmark &  &  & \checkmark &  &  & \textbf{4} \\
            \midrule
            TIFA \citep{hu2023tifa} & \checkmark &  &  &  &  & \checkmark &  & \checkmark & 3 \\
            \midrule
            DSG \citep{cho2023davidsonian} &  &  &  &  &  &  & \checkmark &  & 1 \\
            \midrule
            VQA Score \citep{lin2024evaluating} & \checkmark &  & \checkmark & \checkmark & \checkmark & \checkmark &  & \checkmark & \textbf{6} \\
            \bottomrule
        \end{tabular}
    }
    
    \label{tab:top3_metrics_count}
\end{table*}



\end{document}